\documentclass{article}

\usepackage[preprint]{corl_2026} 

\title{What Actually Matters? A Paired Evaluation of Inference Settings for $\bm{\pi}_{0.5}$ on BEHAVIOR-1K}

\author{
  \textbf{Ruben Hammele$^{1,*}$ \quad
  Samira Huber$^{1,*}$ \quad
  Mohammad Nazeri$^2$ \quad
  Amirreza Payandeh$^2$} \\[0.3em]
  \textbf{Alexandyr T. Card$^2$ \quad
  Aaron Nguyen$^2$ \quad
  Anuj Pokhrel$^2$ \quad
  Daeun Song$^3$} \\[0.3em]
  \textbf{Xuesu Xiao$^2$ \quad
  Sören Pirk$^1$} \\[0.5em]
  $^1$ VCAI Lab, Kiel University, 
  $^2$ RobotiXX Lab, George Mason University\\
  $^3$ AIR Lab, Ewha Womans University \\
}

\begin{document}
\maketitle
\vspace{-1.5em}

\begingroup
\renewcommand{\thefootnote}{*}
\footnotetext{Equal contribution}
\endgroup


\begin{abstract}
    Vision-Language-Action (VLA) models offer a promising paradigm for general-purpose household robotics. However, inference and data-collection strategies that have been deployed in smaller-scale settings do not necessarily generalize to large-scale, diverse task suites. 
    To address this gap, we conduct a systematic, paired evaluation of previously proposed inference and data-collection strategies across all 100 tasks. Our evaluation uses a fine-tuned $\pi_{0.5}$ model based on the 2025 BEHAVIOR Challenge winner, which achieves a Q-score of 0.39 on the public test set.
    These results show that proposed strategies do not transfer uniformly to new environments and tasks.
    Project page, videos, and detailed analysis: \href{https://rubenhammele.github.io/behavior2026}{rubenhammele.github.io/behavior2026}
\end{abstract}

\section{Introduction}

VLA performance depends not only on learned parameters but also on inference and data-collection strategies. Inference strategies are often adopted from prior fine-tuning pipelines, yet their effectiveness beyond the original evaluation settings remains unclear. The inference strategies of the winner of the BEHAVIOR Challenge 2025~\cite{li2024behavior1k}, \citet{larchenko2025taskadaptation}, remain understudied. Other approaches investigate advantage guidance and rollout optimization in the context of garment folding \cite{larchenko2026learningfold}, while industrial reports evaluate $\pi_{0.5}$ fine-tuning and inference settings on a single real-world manufacturing task \cite{dreammachines2026pi05}. The extent to which these strategies generalize across diverse household tasks therefore remains an open question.

We systematically evaluate proposed inference and data-collection strategies using a fine-tuned $\pi_{0.5}$ model across 4 instances of all 100 tasks of BEHAVIOR-1K~\cite{li2024behavior1k}. Halving the number of actions executed per prediction decreases Q-score by 0.16, potentially causing consecutive chunks to pursue different strategies. The inherited $1.3\times$ playback speed-up remains beneficial, likely because it fits more of each task into the time budget. Reducing denoising from twenty steps to one causes no detectable loss, suggesting that a single step already lands close to its mode. For data collection, snapshot-based restarts reduce simulation-step consumption by 29--43\%, since retries resume near the failure point instead of re-simulating already completed stages from the initial state.

\section{Experimental Setup}
\label{sec:setup}

\textbf{Training}
Our policy is initialized from the pre-trained $\pi_{0.5}$ checkpoint introduced by~\citet{larchenko2025taskadaptation}. The policy is fully fine-tuned on the BEHAVIOR-1K dataset. Training then continues in a loop in which the current checkpoint collects rollouts on training instances, and the (partial) successful rollouts are mixed with the demonstrations for further training. Each training frame carries a binary advantage label for RECAP-style conditioning~\citep{intelligence2025pi06}.

\textbf{Benchmark and Statistical Analysis.} We evaluate all 100 BEHAVIOR-1K tasks on public test instances under official constraints~\citep{behavior2026eval}. For each cell, the control is the mean of two independent baseline runs. We assess paired score differences ($\Delta Q$) using task-clustered bootstrap 95\% confidence intervals (CIs), classifying effects as helpful or harmful when the CI excludes zero.

\section{Experiments \& Results}

\subsection{Inference Ablations}

\textbf{Action-Chunk Execution Horizons:} Prior work adopts different action-chunk execution horizons~\citep{black2026pi0,chi2024diffusionpolicy}, while findings on the benefits of longer execution horizons remain task- and policy-dependent~\citep{larchenko2025taskadaptation,bai2025openpicomet}. Consistent with this trade-off, halving the executed prefix from 26 to 13 actions substantially reduced Q-score and success rate while increasing episode length~(Tab.~\ref{tab:inference}). 
\textbf{Playback Speed-Up:} The $1.3\times$ playback speed-up used by~\citet{larchenko2025taskadaptation} also outperformed standard-speed execution, whereas increasing the speed-up to $2\times$ leaned worse without reaching significance.
\textbf{Denoising Steps:} Prior work uses different numbers of flow-matching inference steps, from 4 in GR00T~N1~\citep{nvidia2025gr00tn1} to 10 in $\pi_0$ ~\citep{black2026pi0}. Starting from the 20-step configuration of \citet{larchenko2025taskadaptation}, reducing denoising to 10 or one step produced no detectable performance change.
\textbf{Overlap Inpainting:} Real-Time Chunking (RTC) uses inpainting to maintain chunk continuity during asynchronous inference~\citep{black2025actionchunking}. \citet{larchenko2025taskadaptation} also uses overlap inpainting but does not ablate it. Disabling it had no detectable effect on Q-score.
\textbf{Advantage-Guidance Strength:} Increasing advantage-guidance strength to $\beta=2$ was worse, while $\beta=4$ matched $\beta=1$.
\textbf{Rendering:} Motivated by the gains from high-resolution rendering reported by \citet{bai2025openpicomet}, we compare downsampling with direct rendering at 224 × 224.

\begin{table*}[t]
  \centering

  \begin{minipage}[t]{0.495\textwidth}
    \centering
    \captionof{table}{Inference ablations relative to baseline.}
    \label{tab:inference}
    \resizebox{\linewidth}{!}{%
      \begin{tabular}{llcrrrc}
        \toprule
        Setting & Base. & Tested & $\Delta Q$ & $\Delta$SR
        & $\Delta$Steps & Sig. \\
        \midrule
        Actions/chunk & 26 & 13
        & $-0.16\pm0.03$ & $-11$ & $+752$ & Yes \\
        &  & 30
        & $-0.01\pm0.02$ & $0$ & $-53$ & No \\
        Speed-up & $1.3\times$ & $1\times$
        & $-0.04\pm0.02$ & $-4$ & $+433$ & Yes \\
        & & $2\times$
        & $-0.03\pm0.02$ & $-2$ & $-135$ & No \\
        Denoising & 20 & 10
        & $-0.02\pm0.02$ & $-1$ & $+7$ & No \\
        & & 1
        & $+0.01\pm0.02$ & $1$ & $-23$ & No \\
        Inpainting & On & Off
        & $-0.02\pm0.02$ & $-2$ & $-7$ & No \\
        Guidance $\beta$ & 1 & 2
        & $-0.03\pm0.02$ & $-4$ & $+333$ & Yes \\
        & & 4
        & $-0.00\pm0.02$ & $0$ & $+42$ & No \\
        Gripper Snap & on & off
        & $-0.03\pm0.02$ & $-5$ & $+326$ & No \\
        Rendering & Downsampled & Direct
        & $-0.03\pm0.02$ & $-3$ & $+125$ & No \\
        Sampling & Single & Best-Of-4
        & $-0.01\pm0.02$ & $-2$ & $-1$ & No \\
        \bottomrule
      \end{tabular}%
    }
  \end{minipage}
  \hfill
  \begin{minipage}[t]{0.495\textwidth}
    \centering
    \captionof{table}{Data collection ablations.}
    \label{tab:collection}
    \resizebox{1\linewidth}{!}{%
      \begin{tabular}{lrrr}
  \toprule
  Comparison & $\Delta Q$ & $\Delta$SR & Steps \\
  \midrule
  \multicolumn{4}{l}{\textit{Restart point (Baseline: fresh start)}} \\
  \quad Before first drop & $+0.076\pm0.028$ & $+0.3\pm2.5$ & $-29\%$ \\
  \quad Before last progress & $+0.059\pm0.023$ & $-3.1\pm2.1$ & $-38\%$ \\
  \quad At last progress & $+0.056\pm0.021$ & $-3.4\pm2.3$ & $-43\%$ \\
  \midrule
  \multicolumn{4}{l}{\textit{Modified restore vs.\ full restore}} \\
  \quad Stage belief not restored
  & $-0.05\pm0.02$ & $-5.1\pm2.3$ & -- \\
  \bottomrule
\end{tabular}
    }
  \end{minipage}
\vspace{-0.5em}
\end{table*}

\subsection{Data Collection}
To maximize the number of successful rollouts collected for training, we evaluate strategies to reduce data-collection costs and improve recovery from failed episodes~(Tab.~\ref{tab:collection}).
\textbf{Restart Points:} Prior work replays failure states~\citep{larchenko2026learningfold}, revisits promising states~\citep{DBLP:journals/corr/abs-2004-12919}, and expands start states outward from the goal~\citep{DBLP:journals/corr/FlorensaHWA17}. Among the tested restart points, restarting before the first drop in predicted task progress earned more progress than restarting at the last progress point. \textbf{Simulation Cost:} Restarts reduced simulation cost but did not significantly improve successes per simulation step. \textbf{Stage-Belief Restoration:} Since the policy feeds back its predicted stage~\citep{larchenko2025taskadaptation}, omitting this state during restoration reduced Q-score and task progress, supporting the preservation of policy-side state.

\section{Conclusion}
We have evaluated inference and data-collection strategies across all 100 BEHAVIOR-1K tasks.
Shorter action chunks significantly reduced Q-score and a playback speed of 1.3 improved performance. Fewer denoising steps and disabling overlap inpainting maintained performance, whereas stronger advantage guidance offered no benefit. Snapshot restarts reduced simulation cost but not success, and required restoring the policy’s stage belief. These results suggest that inference and data-collection strategies do not necessarily scale uniformly across task complexity.

\clearpage
\acknowledgments{Thank you to the Stanford BEHAVIOR Challenge organizers, who created this benchmark. This work has taken place in the VCAI Lab at Kiel University and in the RobotiXX Lab at George Mason University. The research was conducted under the RobOdin project funded by the European Union through its Interreg Germany-Denmark funding program. RobotiXX research is supported by National Science Foundation (NSF, 2350352), Army Research Office (ARO, W911NF2320004, W911NF2520011), Army Ground Vehicle Systems Center (GVSC), Google DeepMind (GDM), Microsoft Research (MSR), Clearpath Robotics, FrodoBots Lab, Raytheon Technologies (RTX), Tangenta, 4-VA, Mason Innovation Exchange (MIX), and Walmart. We acknowledge technical support by the Embodied AI Center at Kiel University.}


\bibliography{main}  

\end{document}